# Normalisation of SWIFT Message Counterparties with Feature Extraction and Clustering

Thanasis Schoinas
*Data and Technology*
*Ankura Consulting Group, LLC.*
London, United Kingdom
thanasis.schoinas@ankura.com

Benjamin Guinard
*Data and Technology*
*Ankura Consulting Group, LLC.*
London, United Kingdom
benjamin.guinard@ankura.com

Diba Esbati
*Data and Technology*
*Ankura Consulting Group, LLC.*
London, United Kingdom
diba.esbati@ankura.com

Richard Chalk
*Data and Technology*
*Ankura Consulting Group, LLC.*
London, United Kingdom
richard.chalk@ankura.com

*Abstract* - **Short text clustering is a known use case in the text analytics community. When the structure and content falls in the natural language domain e.g. Twitter posts or instant messages, then natural language techniques can be used, provided texts are of sufficient length to allow for use of (pre)trained models to extract meaningful information, such as part-of-speech or topic annotations. However, natural language models are not suitable for clustering transaction counterparties, as they are found in bank payment messaging systems, such as SWIFT. The manually typed tags are typically physical or legal entity details, which lack sentence structure, while containing all the variations and noise that manual entry introduces. This leaves a gap in an investigator or counter-fraud professional's toolset when looking to augment their knowledge of payment flow originator and beneficiary entities and trace funds and assets. A gap that vendors traditionally try to close with fuzzy matching tools. With these considerations in mind, we are proposing a hybrid string similarity, topic modelling, hierarchical clustering and rule-based pipeline to facilitate clustering of transaction counterparties, also catering for unknown number of expected clusters. We are also devising metrics to supplement the evaluation of the approach, based on the well-known measures of precision and recall. Testing on a real-life labelled dataset demonstrates significantly improved performance over a baseline rule-based ("keyword") approach. At the same time our approach retains most of the interpretability found in rule-based systems, as the former adds an additional level of cluster refinement to the latter. The result is a workflow for standardising populations of entities, with reduced need for manual review. In cases where only a subset of the population needs to be investigated, such as in sanctions investigations, the approach allows for better control of the risks of missing entity variations.**

## I. INTRODUCTION

Named Entity Normalization (NEN) is the name given to the natural language processing technique used to recognize and cluster variations of the same named entity as part of the information retrieval (IR) and named entity recognition (NER) pipelines. With user generated content becoming more and more prevalent, there has been a large amount of research dedicated to the NER and NEN methods from rules based [1] to fuzzy, phonetic or probabilistic approaches [2].

SWIFT (Society for Worldwide Interbank Financial Telecommunication) network messages are structured short text messages, widely used for communication between banks when transferring funds and providing trade services. These are typically made up of a mixture of codes and manually typed payment information that reflect the account details and account holders. Payment information, such as originator, intermediary and beneficiary, can often contain slight variations in name and address details between each payment. This is further complicated with individuals and entities holding accounts at other institutions with further variations. These variations make the use of traditional keyword-based searches and investigations less effective. A method recognising those similarities with minimal human interaction can be beneficial.

A typical use case of named entity normalisation in the investigations industry is sanctions screening. This is based on watch lists issued by authorities and containing Special Designated Nationals (e.g. OFAC SDN), who are individuals and entities located anywhere in the world, usually having interests in multiple jurisdictions and known by different names. Name matching with those lists when investigating a bank's financial systems is crucial, with false positives, i.e. cases where a legitimate person is associated with a sanctioned entity as a result of similarity, being a burden and false negatives, i.e. a transaction associated with a sanctioned entity is not detected by the system, being a risk [3]. Other use cases include anti-money laundering or tax evasion investigations, where standardised entities can make mapping the transactional network and following funds and assets more effective and efficient.

## II. BACKGROUND - RELATED WORK

A classical approach to Named Entity Normalisation comes from using fuzzy matching and string similarity measures such as Levenshtein edit distance; the latter measures the letter by letter similarity between two words [4]. This distance measure can be modified to account for single letter transpositions or substitutions and insertions. In the scope of this paper, a Gestalt Pattern Matching algorithm was used which utilises the longest common subsequence metric and scores similarity between 0 to 1 [5]. These string similarity-based approaches, while effective in accounting for common misspellings, in isolation do not account for major variations in names e.g. Rob Smith and Robert Smith would not be considered the same entity. This is where text analytics approaches, for instance combination with a bag-of-words method, can be found to add value.

On a high level, a typical information retrieval or text analytics pipeline would involve document tokenisation and then various techniques aiming at feature representation, i.e. vectorising the document while also removing "noise", i.e. converting it into a format that is as concise as possible and can be consumed by a supervised or unsupervised learning algorithm to classify or cluster. Various techniques and algorithms can be used for each stage of this pipeline. For instance, vectorisation can include some form of topic modelling, e.g. with Latent Semantic Analysis (LSA), probabilistic LSA [6] or Latent Dirichlet Allocation (LDA) [7], or embedding using a neural network, such as Doc2Vec, or paragraph vectors. The latter outperform bag of words feature representations, by retaining ordering and preserving semantics [8]. Comparisons between the methods, for instance related to the size of the corpus, can be found in various texts, such as [9].

Dictionary based bag-of-words methods such as TF-IDF (Term Frequency-Inverse Document Frequency) have been successfully used in previous applications for topic modelling to aid in document retrieval. The logic behind a TF-IDF model is to generate a matrix of how often terms occur in documents to determine which terms are more representative and "topical" in a document. This also helps filter out noise words that appear frequently across multiple documents which would therefore not be effective keywords [10]. These approaches have successfully been used alongside topic modelling algorithms such as LSA and LDA [11] for NEN in user generated text data [12] and medical applications [13,14]; however very little work exists in the case of little to no available contextual information as is the case with transactional messages.

Once a normalisation pipeline has been applied by using the above measures, the final step is to use the resulting matrices, be it dictionary based, distance based, or a combination, to merge and cluster variations of the entities to reach a final cluster of entities that are assumed by the algorithm to be variations of the same entity. To do this, there are several clustering algorithms available. When dealing with high dimensional data, as is the case with our application, agglomerative or hierarchical clustering, where clusters are merged together using Euclidian distance between the identified features in the feature matrix until a stopping criterion is reached can be applied [15]. The main motivator behind the use of agglomerative clustering is the ability to cluster without knowing the optimal number of clusters (or unique entities) in the dataset prior to the clustering.

Aside from the text variations mentioned earlier, the fairly standard way of inputting entities' details in a SWIFT message makes the use of rule-based elements feasible, when the problem is well defined. For instance, when standardising bank entity details in the above context, the first words are the most important, as they normally contain the person's first and surname, or the company name. A first level, rule-based segmentation can be achieved by these first words. In this trivial segmentation, one can observe false negatives (i.e. reduced recall), introduced by salutations, middle or mistyped names and also false positives (i.e. reduced precision / purity), due to inclusion of potentially separate entities, as indicated by additional names and different addresses. These need to be investigated for relationship with the "perceived main" entity represented by the segment on a one-by-one basis. At the other extreme, if only machine learning is used for segmentation, then variations or mistypes can lead to separating instances of the same entity, introducing false negatives and, conversely, distinct entities sharing elements of their name or address can be clustered together, increasing false positive rates. In an unsupervised learning setting, these problems are further exacerbated by the lack of knowledge about the expected number of clusters.

The existence of a gold dataset facilitates the evaluation of any normalisation or clustering approach, as it enables for the use of external measures using the standard considerations of false positives and negatives, rather than internal measures, dealing solely with the optimisation of criteria such as intra and inter-cluster distance, as for example the Silhouette method. Two typical external measures are Mutual Information and Rand Index [16], with the common theme being penalising the model for both clustering instances of different entities together (FP - false positives) and failing to cluster together instances of the same entity (FN - false negatives). Concepts from set theory, e.g. Jaccard index, are related to the above and are also used in various domains [17,18,19].

III. METHOD

The environment used was Python and the workflow can be separated in the following stages:

*A. Features Extraction*

Each entity short text extracted from a SWIFT tag typically contains name(s) and address details. Along with tokenisation and deduplication and to keep dimensionality from exploding, pre-processing included removal of bank account and any other numeric digits and, finally, special characters. In order to account for commonly used words, e.g. articles, company types (e.g. limited, corporation) etc. instead of removing them, we applied TF-IDF, so those terms would be given a low score if abundant in the dataset. Topic modelling then produced the first family of features. The initial assumption made during topic modelling was that each entity can be viewed as one topic, so a relevant algorithm could reduce dimensionality and add the necessary abstraction to cater for mis-typings and alternative names. However, as the number of entities is very large and typically each entity can have two to three clean text variations on average, the number of topics is expected to be correspondingly high - about half the number of unique data points.

A very popular technique for topic modelling is LDA, with results arguably better than LSA in most cases. We conducted a few initial experiments with this in mind, considering that we could optimise this method to take advantage of the ability to encourage document/topic sparsity through Dirichlet priors.

This would help us apply our design assumption that each "document" (i.e. entity detail short text) should be related to one topic (the main perceived entity itself). However, results (not shown) did not encourage further exploration, as the topics identified were too generic. In addition, one of the strengths of LDA, its ability to generalise well to unknown text, was not deemed important, as investigations are typically performed on datasets collected and frozen at the start, hence overfitting is not an issue. We are also recognising that LDA has complexities, expressed as selection of number of topics and word/topic and topic/document priors requiring optimisation - usually with grid search methods; in addition, it has been expressed in the past that the "degree to which LDA is robust to a poor setting of T (number of topics) is not well-understood" [20]. For these reasons and for the purposes of proving that our general approach works, we resorted to LSA as a baseline, easy to use topic modelling technique, providing the first features family.

The large number of expected topics has also driven the selection of the second family of features. Taking the "worst case scenario", that each entity will have one unique clean text variation and, as such, could be a topic in its own right, we computed a *m x m* matrix, containing similarity scores amongst all *m* unique alphabetised token sets, each presented as one string. Alphabetical sorting was used to further standardise for optimal string similarity figures. Intuitively, variations of the same entity are expected to have a high similarity index, allowing us to treat those strings as topics in a loose sense. Results confirm this theory, as this feature family significantly improved results.

### B. Dimensionality Reduction

Dimensionality reduction was used only in the context of LSA, where the explained variance, as a result of the selected number of components, was tracked and included in the results evaluation.

### C. Clustering

*1) Algorithm:* We performed clustering of the feature vectors using agglomerative clustering, the main driver behind it being the expected number of clusters was unknown. Similarity was judged by Euclidean distance.

*2) Number of Clusters:* The ability to control, with rules-based criteria, the agglomeration stopping points depending on the nature of our problem was also essential in the selection of the clustering algorithm. Given that for the purposes of this paper, we limited our problem to finding variations of the same entity, the rules consisted of allowing two clusters to be merged into one, thus continuing the agglomeration, only when their first or second tokens were of sufficient (75%) similarity, respectively (first-with-first / second-with-second) or interchangeably (first-with-second / second-with-first). These rules, albeit suitable for our use case, can be modified for other use cases. For instance, one may want to allow agglomeration to continue so it can find similarities among entities that share an address rather than a name, which can prove useful in the context of an investigation, as it may indicate a suspicious relationship.

### D. Datasets and Experiments

The starting raw dataset contained about 43k SWIFT tag values as drawn from a real-world case. Deduplication and cleansing, as described earlier, resulted to 2.55k unique lines of entity details, which was further reduced slightly to 2.5k unique alphabetised token sets. The latter was the dataset that went through features extraction and the remaining pipeline.

We performed the following experiments, driven by the feature families used:

1) Lower bounds and Baseline
2) One-Hot / TF-IDF / LSA
3) String Similarity
4) Combined 2 and 3 above
5) Effect of higher order n-grams (bigrams)

The remaining pipeline, i.e. clustering and stopping criteria, was driven by the nature of our use case and common to all the feature families.

The gold dataset was created by manually reviewing the list of 2.55k unique lines. A gold cluster was defined by name and address of an entity, i.e. the same entity across different locations is considered as separate cluster. This produced 1,668 gold clusters.

### E. Measuring performance

We used the Adjusted Mutual Information (AMI) metric to evaluate the models' effectiveness. The selection was based on a few factors: we don't account for class labels, it is widely used and also penalises high number of clusters (that by nature demonstrate high purity). This metric also has many common elements with the traditional document classification metrics used in our industry, precision and recall, but interpreted in a clustering use case. Apart from AMI and to further explore the contribution of precision and recall, we also constructed measures of these as follows: Given an experiment's results, each of the 2.55k unique lines and the gold and machine cluster they belong to, as a list of triplets *(unique line - gold cluster ID - machine cluster ID)*, we calculated the Jaccard indices between each set defined by the intersection of a gold and machine cluster and the sets defined by the respective gold (for recall) and machine (for precision) cluster. The "aggregated" precision and recall for each experiment were the two harmonic means of the individual Jaccard indices lists above. From here on, when mentioning these two metrics, we are referring to the harmonic means of them over all [gold cluster ID - machine cluster ID] combinations in the experiment. Note that, as these two measures do not take account of the size of the clusters, there may be inconsistencies between performance observed based on those and the formal AMI measure. This is the reason why precision and recall as we defined them were used as complementary rather than main measures.

*F. Results*

The results presented in the paragraphs below refer to the use of unigrams only, as these demonstrated the best performance; we reached that conclusion after carrying out a more limited series of experiments with the inclusion of bigrams. Results for the latter are shown at the end of this section.

*1) Establishing bounds and baseline:* Performance bounds were established by taking two extreme scenarios of precision and recall. The baseline was established by a "do-nothing" scenario, where only basic keyword-type rules were used for clustering.

*a) Maximum Recall, "One cluster" - for all entities:* This scenario represents 100% recall, as the single-cluster result contains all entities and their variations. However, it suffers from minimal purity / precision, as all entities are in one cluster. As expected, both precision and AMI were 0.

*b) Maximum Precision, "m clusters" - one per instance:* This scenario is the other extreme of the scenario above. It represents 100% purity and precision, as all single-entity clusters are pure, however the recall for each cluster is the minimum value, as the logic mis-clusters all the other variations of the same entity. As there is a significant number of single entity-clusters in the original dataset, for which the recall will be 100% even in this artificial scenario, the total recall does not limit down to 0 but to a lower bound. This bound was 28% for our dataset. AMI was 0.

*c) Baseline, Keyword-based:* Clusters were created solely on the first two word tokens of the unique line of detail strings. Simply, all entity variations sharing the same first two words formed one cluster. AMI was 67.8%, with precision 56% and recall 80%. The reasonable numbers align with the business logic mentioned earlier, in that the first two tokens are the most important in determining an entity in a SWIFT tag context. This is the baseline result referenced in the rest of this document.

*2) Feature Family 1 - TF-IDF and LSA topic modelling:* Results improved by 13-15% over the baseline with the use of feature extraction and clustering using the first family of features, with the highest increase when using TF-IDF (15%), closely followed by topic modelling (14%), as displayed in Fig. 1. Using one-hot encoding, rather than TF-IDF, improves performance only marginally over the baseline, yielding 68.4% AMI.

The effectiveness of LSA features seems to peak at AMI 77.2%, when the number of components (topics) explain about 90% of the dataset variance. Performance then drops dramatically with increasing explained variance, as the higher number of dimensions is not translated to performance gains. To explain this, we have to remember that, during agglomeration, clusters that don't contain similar first and/or second tokens are not merged, and the branch stops. Consequently, as the number of topics increases, agglomeration tries (and is not allowed) to merge clusters that have short distance due to topics that are related to tokens other than the first two, for example addresses. These terminations drive up the number of resulting clusters and down the overall performance, apart from purity. Note that as we have discussed, purity is increasing with the number of clusters in all clustering cases, but this cannot be accounted as improved performance as this higher number of clusters also increases the risk of false negatives. Finally, even though TF-IDF maintains noise in the form of token variations, it retains all the information without transformations to topics and as such its results are on par with the top LSA model.

By the definition of LSA and by examining the top components, we validate that they primarily relate to words that appear frequently in the dataset. In addition, the related tokens in a component, are terms that cooccur with those terms to a large extent. Both these effects are weaker as we move towards later components that explain less variance. Their effect then becomes detrimental after the 90% cumulative variance. By examining a few of these lower components, we can conclude that these represent noise in our problem. In contrast, the top components indicate relationships between entities, that cause a good performance when only LSA is used and also reinforce these relationships when used in combination with the string matching features; this reinforcement can be attributed to the broader nature of the LSA topics that can surface relationships going beyond the targeted string similarity; for example, there were instances where one variation contained just the company name and another the same company name in a much longer string containing the address. These two instances were far apart in terms of string similarity, but topic features brought them much closer, due to the company name appearing in the same topics for both. On the other hand, geographical relationships identified by any of the feature families (TF-IDF, LSA or Similarity), will be rejected by the rule based agglomeration only if they correspond to different first two tokens in our problem; therefore, for our problem these are not detrimental but, in a different setting, the rule based selection should be adjusted (or removed) to reflect the desirable outcome, for instance entities sharing common address elements.

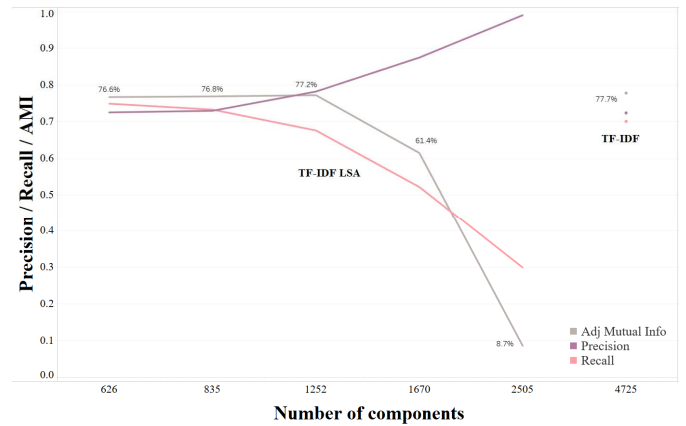

Fig. 1. Relationship of LSA results with number of components (topics). TF-IDF results (with its dimensionality on the x-axis), before Singular Value Decomposition and truncation are also referenced for comparison.

*3) Feature Family 2 - String similarities:* This performed better than the best model of the previous family (81.7% from 77.7% AMI), which can be explained if we consider the string similarity as a measure of the two strings belonging to the same - more targeted - "topic", as mentioned in the features section earlier.

*4) Combining features:* This was the best performing family of experiments, 1.6% better over the similarity features model, proving the logic that we can combine lower (similarity) and higher (topic) level features to achieve maximum performance. Results are displayed in Fig. 2. Overall best results per model type are displayed in Fig. 3. and more detailed results for the - optimally performing - 0.75 string similarity agglomeration stopping rule are displayed in TABLE I. The best performing model overall is the combined string similarity / LSA with 97.5% explained variance. This is marginally better than the string similarity / TF-IDF and seems to peak at higher number of topics compared to the LSA-only models (circa 90% explained variance). Further experiments across different and larger datasets are needed to solidify the results, as they may be dataset specific; due to the high number of single-member clusters in our dataset, the results are probably sensitive to changes even in a few clusters that topic modelling has done a good job at identifying.

In the same context, a consideration about the good results achieved by TF-IDF can be expressed when looking a bit closer into what TF-IDF does: as it is applied to unique deduplicated token sets, TF will be 1 or 0 in all cases. On the other hand, IDF is where the contribution of common words is controlled, by assigning lower weights to those. This appears to have a positive effect as shown from the results, especially given our lack of a stop words list. However, this characteristic is not beneficial in cases of spelling mistakes that exist in very few instances, as TF-IDF will assign a high value to them, causing the distance between the correct entity and the one with the misspelling to increase, thus increasing the risk of the two instances not being clustered together.

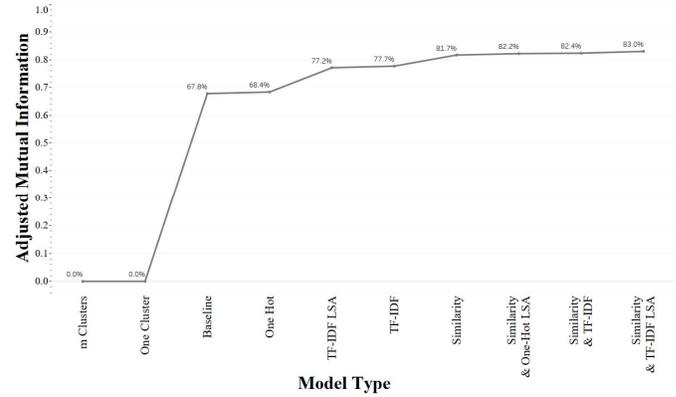

Fig. 3. Best achieved results per modelling approach (unigrams-only models).

Another side-effect is that TF-IDF will penalise tokens signifying common geographical locations, e.g. large cities, as these will be more abundant in the dataset. This is exactly the effect that LSA or - potentially even better - LSA on One-Hot, rather than TF-IDF, term-document matrix should be able to alleviate. Results with LSA applied on One-Hot are so far showing that TF-IDFs contribution towards unwanted common words is stronger than its shortcomings for wanted common words or rare spelling mistakes. These scenarios need to be tested further on various datasets; nevertheless, our solution could benefit from a user-defined unwanted common words list to be removed in advance, so the need for TF-IDF no longer exists.

*5) Effect of higher order n-grams (bigrams):* Including higher order n-grams does not seem to have the potential to improve performance on our dataset. Bigrams were created from the unique lines of entity details - before the final alphabetisation - to examine whether maintaining some of the token ordering can add value, presuming that this could be useful in cases of multi-token entity names or geographical references. When bigrams where included in the TF-IDF computations, results were inferior to those obtained with unigrams only, as shown in Fig. 4 and Fig. 5, indicating that adding bigrams increases noise and a standard bag of unigrams can be sufficient for the special nature of our problem.

We also observe a less steep drop in performance with the increase of components, which is expected, as the use of bigrams introduces higher dimensionality to the feature set, therefore the same number of topics explains less of the dataset variance compared to unigrams only. Note that, due to the number of components being restricted by the limited number of available datapoints, we cannot conclude with certainty whether the bigram results would be the same if we had a larger dataset, as we can see that the best performing combined bigram model is the one with the highest possible number of topics (100% explained variance, $m$-1 number of topics, where $m$ is the number of data points).

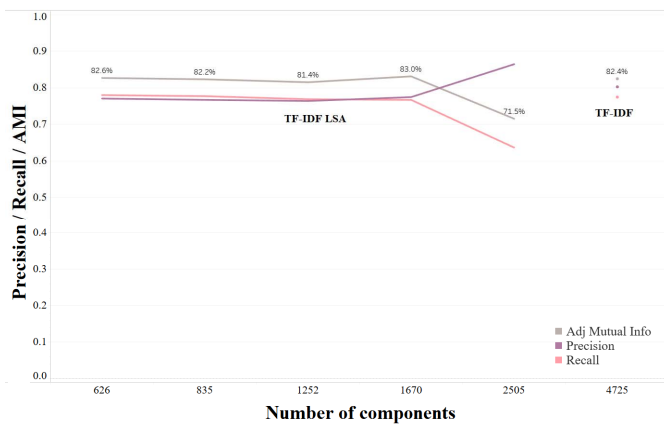

Fig. 2. Performance of combined String similarity - LSA features for various numbers of components (string similarity dimensions are excluded). TF-IDF (with its dimensionality on the x-axis) is also presented for reference.

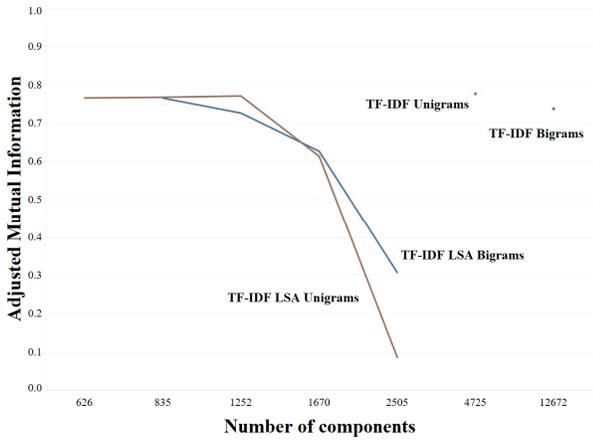

Fig. 4. Bigrams/unigrams versus unigrams, in relation to LSA results and number of components. TF-IDF results are also referenced for comparison.

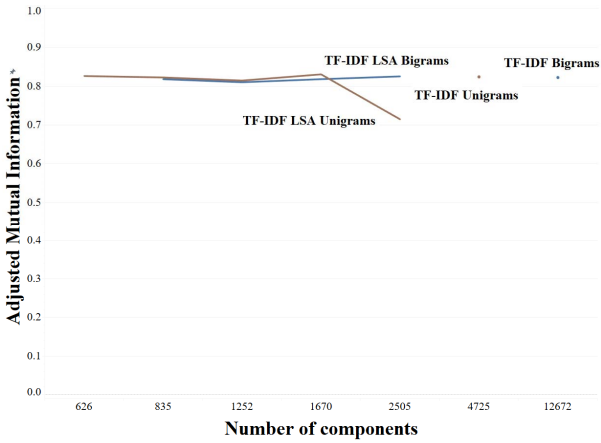

Fig. 5. Bigrams/unigrams versus unigrams. Performance of combined String similarity - LSA features for various numbers of components (string similarity dimensions are excluded). TF-IDF results are also presented for reference.

Even though the vocabulary size will tend to increase with the number of datapoints, we can certainly expect situations where the number of points will be higher than the number of dimensions, as would be the case in high overlap of names or geographies among the unique token sets. In such a situation, a larger number of components might drive a bigram model performance higher than that of the unigram model.

*G. Scaling*

The most demanding and computationally intensive stage is the creation of the string similarity feature matrix, as it has $O(m^2 \times n^3)$ complexity, where $m$ the number of datapoints and $n$ the string length. Selection of a less expensive string-matching algorithm could improve scaling; however, it needs to be tested for trade-off in accuracy. This work is necessary though, as an experiment on a large bank dataset over various SWIFT tag types would require hundreds of billions of string comparisons.

An interim solution, for environments of limited capacity, is to break down the population - e.g. per starting letter of the alphabet - and then combine the results and potentially rerun on the normalised chunks, until all normalised entities can be processed in one chunk.

Finally, it is worth noting that, as is the case with all string pair comparison methods, the similarity matrix calculation can be easily chunked and parallelized for execution in a distributed environment. This can cut down runtime by orders of magnitude.

## IV. CONCLUSIONS

We have created a customizable workflow of named entities normalisation and demonstrated its application on short texts, such as SWIFT free text tags. A few observations were made along the way:

TABLE I.    BEST RESULTS FOR OPTIMAL (0.75) STRING SIMILARITY THRESHOLD AT RULE-BASED AGGLOMERATION STOPPING (WHERE APPLICABLE)

| Model Type | Settings and Results | | | | | |
|---|---|---|---|---|---|---|
| | *Explained Variance* | *Number of Dimensions* | *Number of Clusters* | *Recall* | *Precision* | *Adjusted Mutual Information* |
| Baseline | - | - | 1,526 | 80.1% | 56.1% | 67.8% |
| m Clusters | - | - | 2,552 | 27.9% | 100% | 0.0% |
| One Cluster | - | - | 1 | 100% | 0.0% | 0.0% |
| One-Hot | - | 4,753 | 1,817 | 55.6% | 86.8% | 68.4% |
| Similarity | - | 2,505 | 1,701 | 76.9% | 80.8% | 81.7% |
| Combined (One-Hot LSA) | 83.9% | 3,131 | 1,680 | 77.2% | 80.8% | 82.2% |
| TF-IDF | - | 4,725 | 1,618 | 70.0% | 72.3% | 77.7% |
| TF-IDF Bigrams | - | 12,672 | 1,732 | 63.7% | 78.4% | 73.8% |
| TF-IDF LSA | 89.8% | 1,252 | 1,622 | 67.5% | 78.1% | 77.2% |
| TF-IDF LSA Bigrams | 67.5% | 835 | 1,589 | 68.5% | 73.8% | 76.6% |
| Combined (TF-IDF) | - | 7,230 | 1,686 | 77.4% | 80.2% | 82.4% |
| Combined (TF-IDF Bigrams) | - | 15,177 | 1,692 | 77.1% | 80.5% | 82.2% |
| Combined (TF-IDF LSA) | 97.5% | 4,175 | 1,673 | 76.6% | 77.4% | 83.0% |
| Combined (TF-IDF LSA Bigrams) | 100% | 5,010 | 1,683 | 73.8% | 79.3% | 82.5% |

- A matrix of string similarity scores between each pair of entity variations was the most effective feature set, reaching 81.7% Adjusted Mutual Information score when used as the sole feature family. This can be considered as a form of topic model, as pairs of variations with a high similarity score can be seen as belonging to the same "topic", i.e. the same latent entity where the variations stem from.

- The above can be combined with traditional topic modelling features, such as LSA, where variations of the same entity with lower similarity scores can be clustered together based on topic, rather than pure string similarity features. This combination improved performance by circa another 1.5% AMI.

- LDA, a very capable method for natural language topic modelling, was not suited to our problem, as it produces sparse and over inclusive topics. Perhaps with a more informed selection of priors it could have performed better.

- The simpler, non-sparse and algebraic LSA produced more manageable results and ability to have control over them, by modifying only the number of components and assessing explained variance.

- Using bigrams, as a feasibility experiment on the use of higher order n-grams, has not shown improvements. However, it is worth noting that our dataset was rather small and contained more dimensions than number of datapoints, especially more so in the case of bigrams, which limited the available components. Even though dimensions would still increase in larger sets, we would expect increasingly high overlap of names or geographies among the unique token sets. In those cases, bigrams performance could potentially improve, combined with suitable dimensionality reduction techniques.

- Clustering is carried out in an agglomerative manner, allowing for flexible number of expected clusters; rule-based agglomeration stopping can be used to control the number, as well the nature of the clusters.

The benefits gained by the use of the approach can be summarised as:

- The reduction of entity groups reviewed, especially for clusters with large number of variations, can be significant.

- Entities and their variations are grouped into reviewable clusters of entity detail strings. The conservative rules-based final stage minimises redundant work to manually split clusters that should not have been joined together (precision), or conversely, controlling the risk of missing entity variations from clusters under review (recall).

- Flexibility in terms of altering the value or modifying the logic of the rule-based similarity threshold, to allow for other ways of stopping the agglomeration and reveal alternative entity relationships; for instance, different entities sharing the same address might, in the context of an investigation, indicate a suspicious relationship.

## V. Future Work

A number of commercial applications exist, performing the task predominantly with Fuzzy matching and other methods. We are also starting to see products that exploit machine learning, so it would be worth performing a comparison, in terms of both functionality and performance on our datasets.

From a scaling perspective, distributing the computation of the similarity matrix can greatly speed up the extraction of this feature family in very large datasets, involving several different SWIFT tag types and large-sized banks' volumes of transactions.

In the same context, extraction of the first family of features, via the use of topic modelling for vectorisation, can benefit from the use of more modern methods, exploiting neural networks and the use of GPUs. With both accuracy and scalability in mind, word and, in particular, paragraph embeddings [21, 22] could potentially substitute LSA. Going one step further, another likely candidate is "lda2vec" that can learn both document and topic vectors at the same time [23], as long as the initialisation of word vectors is done in the right manner. Finally, higher order n-grams could be an interesting area of further work, especially for larger datasets that would allow for better experimentation with dimensionality reduction via decompositions or embeddings.